\documentclass[11pt,a4paper]{article}
\usepackage{authblk}
\usepackage[hyperref]{acl2018}
\usepackage{times}
\usepackage{latexsym}
\usepackage{booktabs} 
\usepackage{graphicx}

\usepackage{url} 

\aclfinalcopy 
\def\aclpaperid{403} 

\title{Fine-grained evaluation of Quality Estimation for Machine translation based on a linguistically-motivated Test Suite}

\author[*]{Eleftherios Avramidis}
\author[*]{Vivien Macketanz}
\author[**]{Arle Lommel}
\author[*]{Hans Uszkoreit}
\affil[*]{German Research Center for Artificial Intelligence (DFKI), Berlin,
Germany  \protect\\ \texttt{firstname.lastname@dfki.de}}
\affil[**]{Common Sense Advisory (CSA Research), Massachusetts, USA
\protect\\
\texttt{alommel@csa-research.com}}

\date{} 

\begin{document}
\maketitle
\begin{abstract}
We present an alternative method of evaluating Quality Estimation systems, which
is based on a linguistically-motivated Test Suite. 
We create a test-set consisting of 14 linguistic error categories
and we gather for each of them a set of samples with both correct and
erroneous translations. 
Then, we measure the performance of 5 Quality Estimation systems by checking
their ability to distinguish between the correct and the erroneous
translations.
The detailed results are much more informative about the ability of each system.
The fact that different Quality Estimation systems perform differently at
various phenomena confirms the usefulness of the Test Suite. 
  
\end{abstract}

\section{Introduction}
\label{sec:intro}

The evaluation of empirical Natural Language Processing (NLP) systems is a
necessary task during research for new methods and ideas.
The evaluation task is the last one to come after the development process and
aims to indicate the overall performance of the newly built system and compare
it against previous versions or other systems.
Additionally, it also allows for conclusions related to the decisions taken for
the development parameters and provides hints for improvement.
Defining evaluation methods that satisfy the original development requirements
is an ongoing field of research. 

Automatic evaluation in sub-fields of Machine Translation (MT) has been mostly
performed on given textual hypothesis sets, where the performance of the system
is measured against gold-standard reference sets with one or more metrics
\cite{bojar-EtAl:2017:WMT1}.
Despite the extensive research on various automatic metrics and scoring methods,
little attention has been paid to the actual content of the test-sets and how
these can be adequate for judging the output from a linguistic perspective.
The text of most test-sets so far has been drawn from various random sources and
the only characteristic that is controlled and reported is the generic domain of
the text.

In this paper we make an effort to demonstrate the value of using a
linguistically-motivated controlled test-set (also known as a \emph{Test Suite})
for evaluation instead of generic test-sets.
We will focus on the sub-field of sentence-level Quality Estimation (QE) on MT
and see how the evaluation of QE on a Test Suite can provide useful information
concerning particular linguistic phenomena.

\section{Related work}
\label{sec:related}

There have been few efforts to use a broadly-defined Test Suite for the
evaluation of MT, the first of them being during the early steps of the
technology~\cite{King1990}.
Although the topic has been recently revived \cite{Isabelle2017,Burchardt2017},
all relevant research so far applies only to the evaluation of MT output and
not of QE predictions.

Similar to MT output, predictions of sentence-level QE have also been evaluated
on test-sets consisting of randomly drawn texts and a single metric has been
used to measure the performance over the entire
text~\cite[e.g.][]{bojar-EtAl:2017:WMT1}.
There has been criticism on the way the test-sets of the shared tasks have been
formed with regards to the distribution of inputs \citep{Anil2013}, e.g. when
they demonstrate a dataset shift \citep[][]{Gretton2009}. 
Additionally, although there has been a lot of effort to infuse linguistically
motivated features in QE \cite{felice-specia:2012:WMT}, there has been no effort
to evaluate their predictions from a linguistic perspective.
To the best of our knowledge there has been no use of a Test Suite in order to
evaluate sentence-level QE, or to inspect the predictions with regards to
linguistic categories or specific error types. 
 
\begin{figure*}
\centering
  \includegraphics[width=0.95\textwidth]{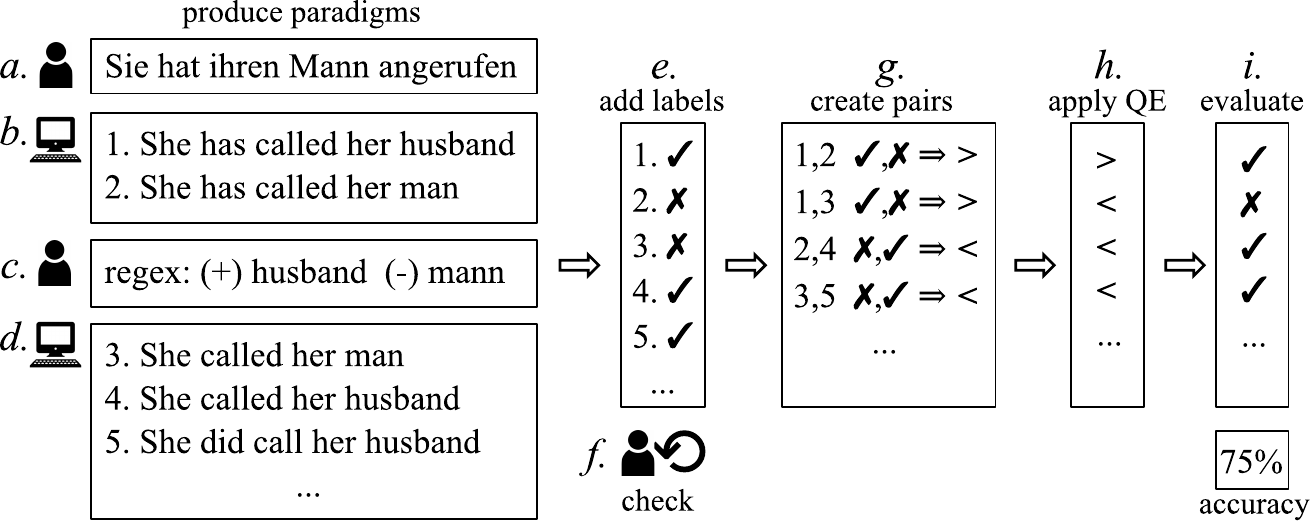}
    \caption{Example for the processing of test items for the lexical ambiguity of word ``Mann''}
  \label{fig:example} 
\end{figure*}

\section{Method}
\label{sec:method}

The evaluation of QE presented in this paper is based on these steps:
(1) construction of the Test Suite with respect to linguistic categories; (2)
selection of suitable Test Suite sentences; and (3) analysis of the Test Suite by
existing QE systems and statistical evaluation of the predictions. These steps
are analysed below, whereas a simplified example is given in Figure \ref{fig:example}.

\subsection{Construction of the Test Suite}

The Test Suite has been developed by a professional linguist, supported by
professional translators. First, the linguist gathers or creates error-specific
paradigms (Figure~\ref{fig:example}, stage~a), i.e. sentences whose translation
has demonstrated or is suspected to demonstrate systematic errors by known MT
engines.
The aim is to have a representative amount of paradigms per error type and the
paradigms are as short as possible in order to focus solely on one phenomenon
under examination.
The error types are defined based on linguistic categories inspired by the MQM
error typology~\cite{pub7426} and extend the error types presented in
\newcite{Burchardt2017}, with additional fine-grained analysis of
sub-categories.
The main categories for German-English can be seen in
Table~\ref{tab:categories}.

Second, the paradigms are given to several MT systems (Figure~\ref{fig:example},
stage~b) to check whether they are able to translate them properly , with the
aim to acquire a ``pass'' or a ``fail'' label accordingly.
In an effort to accelerate the acquisition of these labels, we follow a
semi-automatic annotation method using regular expressions.
The regular expressions allow a faster automatic labelling that focuses on
particular tokens expected to demonstrate the issue, unaffected from alternative
sentence formulations.
For each gathered source sentence the linguist specifies regular expressions
(Figure~\ref{fig:example}, stage~c) that focus on the particular issue: one
positive regular expression that matches a successful translation and gives a
``pass'' label and an optional negative regular expression that matches an
erroneous translation and gives a ``fail'' (for phenomena such as ambiguity and
false friends).
The regular expressions, developed and tested on the first translation outputs,
are afterwards applied to all the alternative translation outputs (stage~d) to
acquire the automatic labels (stage~e).
Further modifications to the regular expressions were applied, if they did not
properly match the new translation outputs.
The automatically assigned labels were controlled in the end by a professional
translator and native speaker of the target language (stage f).
For the purposes of this analysis, we also assume that every sentence paradigm
only demonstrates the error type that it has been chosen for and no other major
errors occur.

\subsection{Selection of suitable Test Suite sentences}
\label{sec:method:sentence_selection}

The next step is to transform the results so that they can be evaluated by
existing sentence-level QE methods, since the Test Suite provides binary
pass/fail values for the errors, whereas most sentence-level QE methods predict
a continuous score.
For this purpose, we transform the problem to a problem of predicting
comparisons.
We deconstruct the alternative translations of every source sentence into
pairwise comparisons, and we only keep the pairs that contain one successful and
one failing translation (Figure~\ref{fig:example}, stage~g).
Sentence-level QE systems will be given every pair of MT outputs and requested
to predict a comparison, i.e. which of the two outputs is better (stage~h).
Finally, the QE systems are evaluated based on their capability to properly
compare the erroneous with the correct outputs (stage~i).
The performance of the QE systems will be therefore expressed in terms of the
accuracy over the pairwise choices.

\section{Experiment}
\label{sec:experiment}

\subsection{Data and systems}
\label{sec:experiment:data}

\begin{table}
\centering
\begin{tabular}{lr}
\toprule
MT type  & proportion \\
\midrule
neural       & 64.7\% \\
phrase-based & 26.8\% \\
both (same output) & 8.5\% \\
\bottomrule
\end{tabular}
\caption{MT type for the translations participating in the final pairwise
test-set}
\label{tab:mt_types}
\end{table}

The current Test Suite contains about 5,500 source sentences and their rules
with regular expressions for translating German to English.
These rules have been applied for evaluating 10,800 unique MT outputs (MT
outputs with the exact same text have been merged together).
These outputs have been produced by three online commercial systems (2
state-of-the-art neural MT systems and one phrase-based), plus the open-source
neural system by \newcite{sennrich-EtAl:2017:WMT}.
After creating pairs of alternative MT outputs that have a different label
(Section~\ref{sec:method:sentence_selection}) the final test-set contains 3,230
pairwise comparisons based on the translations of 1,582 source sentences.
The MT types of the translations participating in the final test-set can be seen
in Table~\ref{tab:mt_types}.

For this comparative study we evaluate existing QE systems that were freely
available to train and use.
In particular we evaluate the baseline the following 6 systems:
\begin{itemize}
\item \textbf{B17:} The baseline of the shared task on sentence-level
QE~\cite{bojar-EtAl:2017:WMT1} based on 17 black-box features
and trained with Support Vector Regression (SVR) to predict continuous HTER
values
\item \textbf{B13:} the winning system of the shared task on
QE ranking~\cite{Bojar2013,Avramidis2013} based on 10 features, trained
with Logistic Regression with Stepwise Feature Selection in order to perform ranking.
Despite being old, this system was chosen as it is the latest paradigm of
Comparative QE that has been extensively compared with competitive methods in a
shared task
\item \textbf{A17:} three variations of the state-of-the-art research 
on Comparative QE \cite{pub9044}, all three trained with a Gradient Boosting 
classifier. The \emph{basic} system has the same feature set as B13,
the \emph{full} system contains a wide variety of 139 features and the
\emph{RFECV} contains the 25 highest ranked features from the full
feature set, after running Recursive Feature Elimination with an SVR 
kernel. 
\end{itemize}  
The implementation was based on the open-source tools Quest~\cite{pub7042} and
Qualitative~\cite{pub8768}.

\subsection{Results}
\label{sec:experiment:results}

\begin{table*}
\centering
\begin{tabular}{lrccccc}
\toprule
	&	& B17	& B13 &  \multicolumn{1}{l}{$\lceil$}   & A17 & \multicolumn{1}{r}{$\rceil$}  \\
	&amount	&  baseline	&  winning  & basic	& RFECV & full\\
\midrule

Ambiguity	              		&89	&58.4	&64.0	&\textbf{73.0}	&69.7	&62.9\\
Composition	             		&75	&58.7	&77.3	&\textbf{80.0}	&72.0	&77.3\\
Coordination \& ellipsis 		&78	&53.8	&\textbf{73.1}	&71.8	&71.8	&70.5\\
False friends	         		&52	&38.5	&32.7	&\textbf{48.1}	&38.5	&42.3\\
Function word	         		&126	&33.3	&\textbf{38.9}	&35.7	&32.5	&34.9\\
Long distance dep. \& interrogatives	&266	&52.3	&63.9	&60.2	&63.9 &\textbf{65.8}\\
Multi-word expressions			 &43	&32.6	&\textbf{44.2}	&32.6	&39.5	&39.5\\
Named entity \& terminology		&55	&50.9	&54.5	&56.4	&58.2	&\textbf{60.0}\\
Negation						&13	&38.5	&53.8	&\textbf{76.9}	&\textbf{76.9}	&\textbf{76.9}\\
Non-verbal agreement			&45	&40.0	&\textbf{57.8}	&53.3	&\textbf{57.8}	&53.3\\
Punctuation						&138	&11.6	&29.7	&\textbf{32.6}	&28.3	&27.5\\
Subordination					&46	&41.3	&43.5	&\textbf{47.8}	&45.7	&\textbf{47.8}\\
Verb tense/aspect/mood/type			&2137	&56.6	&\textbf{59.4}	&55.5	&57.3	&57.7\\
Verb valency					&67	&50.7	&55.2	&50.7	&58.2	&\textbf{62.7}\\
\midrule
Total							&3230	&52.1	& \textbf{57.5}	&55.0	&56.1	&56.7\\
weighed							&	&44.1	&53.4	&55.3	&55.0	&\textbf{55.6}\\
\bottomrule
\end{tabular}

\caption{QE accuracy (\%) per error category}
\label{tab:categories}
\end{table*}

Here we present the evaluation of the QE systems when applied on the Test Suite.
The accuracy achieved by each of the 6 QE systems for the 14 error categories
can be seen in Table~\ref{tab:categories}.

First, it can be noted that the \textbf{quantity of evaluated samples} varies a
lot and, although the original aim was to have about 100 samples per category,
most of the neural outputs succeeded in the translations of the issues and
therefore were not included in the test-set with the ``pass/fail" comparisons.
Obviously, conclusions for those error categories with few samples
cannot be guaranteed.

Second, one can see that the \textbf{average scores} range between 52.1\% and
57.5\% (achieved by B13) which are nevertheless relatively low.
This may be explained by the fact that all QE systems have been developed in the
previous years with the focus on ``real text" test-sets.
The Test Suite on the contrary is not representative of a real scenario and has
a different distribution than the one expected from real data.
Additionally, many of the linguistic phenomena of the Test Suite may have few or
no occurrences on the development data of the QE systems.
Finally, all QE systems have been developed in the previous years with the focus
on rule-based or phrase-based statistical MT and therefore their performance on
MT output primarily from neural systems is unpredictable.

We also report scores averaged not out of the total amount of the samples, but
instead giving equal importance to each error category.
These scores indicate a different winner: the full system of A17.
However, due to the distributional shift of the Test Suite,
there is limited value in drawing conclusions from average scores, since the aim
of the Test Suite is to provide a qualitative overview of the particular linguistic
phenomena.

\begin{table*}
\centering
\begin{tabular}{lrccccc}
\toprule 
	&	& B17	& B13 &  \multicolumn{1}{l}{$\lceil$}   & A17 & \multicolumn{1}{r}{$\rceil$}  \\
	&amount	&  baseline	&  winning  & basic	& RFECV & full\\
\midrule
 future I	                &297	&\textbf{58.9}	&\textbf{58.9}	&52.5	&50.5	&51.5\\
 future I subjunctive II	&249	&\textbf{62.7}	&52.6	&45.0	&51.4	&53.0\\
 future II	                &158	&39.2	&56.3	&\textbf{60.1}	&58.2	&53.2\\
 future II subjunctive II	&168	&32.7	&\textbf{78.0}	&74.4	&68.5	&75.6\\
 perfect	                &294	&55.4	&\textbf{56.8}	&49.3	&55.8	&54.8\\
 pluperfect					&282	&\textbf{72.7}	&65.6	&64.9	&69.9	&68.1\\
 pluperfect subjunctive II	&159	&52.2	&53.5	&\textbf{55.3}	&52.8	&\textbf{55.3}\\
 present					&286	&\textbf{58.0}	&54.9	&51.4	&51.0	&52.8\\
 preterite					&105	&61.0	&\textbf{68.6}	&53.3	&67.6	&\textbf{68.6}\\
 preterite subjunctive II	&88     &\textbf{62.5}	&61.4	&58.0	&53.4	&55.7\\
\bottomrule
\end{tabular}
\caption{QE accuracy (\%) on error types related to verb tenses}
\label{tab:tenses}
\end{table*}

\begin{table*}
\centering
\begin{tabular}{lrccccc}
\toprule
	&	& B17	& B13 &  \multicolumn{1}{l}{$\lceil$}   & A17 & \multicolumn{1}{r}{$\rceil$}  \\
	&amount	&  baseline	&  winning  & basic	& RFECV & full\\
\midrule
Ditransitive 	&275	&46.9	&57.8	&55.6	&56.4	&\textbf{60.0}\\
Intransitive 	&171	&42.1	&\textbf{69.6}	&57.3	&59.1	&64.3\\
Modal 	&473	&63.4	&67.2	&57.9	&66.6	&\textbf{67.2}\\
Modal negated 	&657	&\textbf{70.3}	&49.9	&47.2	&46.0	&46.3\\
Reflexive 	&376	&44.7	&61.2	&61.2	&\textbf{62.2}	&58.5\\
Transitive 	&134	&39.6	&68.7	&69.4	&64.9	&\textbf{68.7}\\
\bottomrule
\end{tabular}
\caption{QE accuracy (\%) on error types related to verb types}
\label{tab:verbmoods}
\end{table*}

When it comes to \textbf{particular error categories}, the three systems B13,
A17-basic and A17-full seem to be complementary, achieving the highest score for
5 different error categories each.
The systems B17 and A17-RFECV lack a lot in their performance.
The highest category score is achieved for the phenomenon of \emph{Composition}
(compounds and phrasal verbs) by A17-basic, followed by \emph{negation} (albeit
with very few samples) at 76.9\%.
A17-basic is also very strong in \emph{ambiguity}, achieving 73\%.
The 4 systems B13 and A17 perform much better concerning \emph{long-distance
relationships}, which may be attributed to the parsing and grammatical features
they contain, as opposed to the B17 which does not include parsing.
Finally, A17-full does better with \emph{named entities} and \emph{terminology},
possibly because its features include alignment scores from IBM model 1.

We notice that \textbf{verb tenses, aspects, moods and types} comprise a major
error category which contains more than 2,000 samples.
This enables us to look into the subcategories related to the verbs.
The performance of the systems for different tenses can be seen in
Table~\ref{tab:tenses}, where B17 and B13 are the winning systems for 5
categories each.
The tense with the best performance is the \emph{future II subjunctive II} with
a 78\% accuracy by B13.
Despite its success in the broad spectrum of error categories, A17-full performs
relatively poorly on verb tenses.

Finally, Table~\ref{tab:verbmoods} contains the accuracy scores for
\textbf{verb types}. 
A17-full does much better on verb types, with the exception of the \emph{negated
modal} which gets a surprising 70.3\% accuracy from B17.

\section{Conclusion and further work}
\label{sec:conclusion}

In this paper we demonstrated the possibility of performing evaluation of QE by
testing its predictions on a fine-grained error typology from a Test Suite.
In this way, rather than judging QE systems based on a single score, we were able to see
how each QE system performs with respect to particular error categories.
The results indicate that no system is a clear winner, with three out of the 5
QE systems to have complementary results for all the error categories.
The fact that different QE systems with similar overall scores perform
differently at various phenomena confirms the usefulness of the Test
Suite for understanding their comparative performance.

Such linguistically-motivated evaluation can be useful in many aspects.
The development or improvement of QE systems may use the results about the found errors in order to introduce new related features.
The development may also be aided by testing these improvements on an isolated
development set.

Further work should include the expansion of the Test Suite with more samples in
the less-populated categories and support for other language pairs.
Finally, we would ideally like to broaden the comparison among QE systems,
by including other state-of-the-art ones that unfortunately were not freely
available to test.

\section*{Acknowledgments} Part of this work has received funding from the EU
Horizon 2020 research and innovation program QT21 under grant agreement N$^{o}$
645452.

\bibliographystyle{acl_natbib}
\bibliography{library} 

\appendix

\label{sec:supplemental}

\end{document}